\documentclass[CoRe Paper 2021={Social Media for Crisis Management},anonymous=false]{iscram}
\usepackage{algpseudocode}% -*- coding: utf-8; -*-
\usepackage[utf8]{inputenc}
\usepackage{algorithm}
\usepackage{subcaption}
\usepackage{lipsum}
\usepackage{tikz}
\usepackage{fancyvrb}
\usepackage{listings}
\usepackage{enumitem}
\usepackage{tcolorbox}
\usepackage{multicol}
\usepackage{siunitx}
\usepackage{multirow}

\usepackage{MnSymbol}
\lstdefinestyle{common}{
  xleftmargin=.5em,
  xrightmargin=.5em,
  frame=single,framesep=.5em,framerule=0pt,
  fancyvrb=true,
  basicstyle=\ttfamily,
  keywordstyle=\color{cyan!50!blue!75!black}\bfseries,
  commentstyle=\color{red!50!black}\itshape,
  stringstyle=\ttfamily\color{green!50!black},
  numbers=none,
  showspaces=false,
  showstringspaces=false,
  fontadjust=true,
  keepspaces=true,
  flexiblecolumns=true,
  emphstyle=\color{red},
}
\lstdefinestyle{TeX}{
  style=common,
  backgroundcolor=\color{blue!5},
  aboveskip=5pt,
  belowskip=5pt,
  language=[LaTeX]TeX,
  moretexcs={
    abstract, addbibresource, iscramset, keywords, mainmatter,
    maketitle, printbibliography, subsection, subsubsection, url,
    urldef, href, includegraphics, ldots, parencite, citeauthor,
    citeyear, citetitle, midrule, toprule, bottomrule
  },
  fancyvrb=true,
}
\lstdefinestyle{console}{
  style=common,
  backgroundcolor=\color{gray!10},
  aboveskip=5pt,
  belowskip=5pt,
}

\newlist{options}{description}{1}
\setlist[options]{%
  beginpenalty=10000,%
  itemsep=.5\parskip plus .3\parskip minus .2\parskip,
  parsep=.5\parskip plus .3\parskip minus .2\parskip,
  topsep=.5\parskip plus .3\parskip minus .2\parskip,
  partopsep=.5\parskip plus .3\parskip minus .2\parskip,
  style=nextline,labelindent=1em,%
  font=\normalfont\ttfamily}

\colorlet{macro color}{cyan!50!blue!75!black}
\colorlet{option color}{red!50!black}
\colorlet{generic color}{green!40!black}

\newtcolorbox{pseudoTeX}{colback=blue!5,colframe=blue!5,before=\nobreak}
\let\LaTeXorig\LaTeX
\renewcommand\LaTeX{\bgroup\fontfamily{lmr}\selectfont\upshape\LaTeXorig\egroup}

\newcommand\ric[1]{\textcolor{black}{#1}}

\newcommand{\zijunnew}[1]{\textcolor{black}{#1}}

\iscramset{
  title={
CrisisViT: A Robust Vision Transformer for Crisis Image Classification
  },
  short title={CrisisViT},
  author={
    short name={Zijun Long},
    full name=Zijun Long,
    affiliation={University of Glasgow\\ \href{mailto:z.long.2@research.gla.ac.uk}\url{z.long.2@research.gla.ac.uk}},
  },
  author={
    short name={Richard McCreadie},
    full name=Richard McCreadie,
    affiliation={University of Glasgow\\ \href{mailto:richard.mccreadie@glasgow.ac.uk}\url{richard.mccreadie@glasgow.ac.uk}},
  },
    author={
    short name={Muhammad Imran},
    full name=Muhammad Imran,
    affiliation={Qatar Computing Research Institute\\ Hamad Bin Khalifa University\\ \href{mailto:mimran@hbku.edu.qa}\url{mimran@hbku.edu.qa}},
  },
}
\usepackage{tikzpagenodes}
\usetikzlibrary{fit}

\begin{document}

\maketitle

%\makeatletter
%{\centering\large\iscram@version{}\\\iscram@date\par}
\makeatother

\abstract{

\zijunnew{In times of emergency, crisis response agencies need to quickly and accurately assess the situation on the ground in order to deploy relevant services and resources. However, authorities often have to make decisions based on limited information, as data on affected regions can be scarce until local response services can provide first-hand reports. Fortunately, the widespread availability of smartphones with high-quality cameras has made citizen journalism through social media a valuable source of information for crisis responders. However, analyzing the large volume of images posted by citizens requires more time and effort than is typically available. To address this issue, this paper proposes the use of state-of-the-art deep neural models for automatic image classification/tagging, specifically by adapting transformer-based architectures for crisis image classification (CrisisViT). We leverage the new Incidents1M crisis image dataset to develop a range of new transformer-based image classification models. Through experimentation over the standard Crisis image benchmark dataset, we demonstrate that the CrisisViT models significantly outperform previous approaches in emergency type, image relevance, humanitarian category, and damage severity classification. Additionally, we show that the new Incidents1M dataset can further augment the CrisisViT models resulting in an additional 1.25\% absolute accuracy gain.}

}


\section{Introduction}

\ric{Crisis events, such as floods, fires and COVID-19, generate significant attention from both news media and the general public, leading to related content being posted to a wide variety of social media platforms, such as Twitter or Facebook. Previous studies~\parencite{RN82, RN83, RN84} have demonstrated the importance of using social media as a way to acquire information during a crisis event. However, the limited time available to emergency responders in combination with the large volume of posts made on these platforms necessitates automated tooling to extract only the actionable portions of that content~\parencite{mccreadie2020incident, 28widener2014using}. Indeed, over the last decade there have been a wide range of works examining how machine learning can be used to aid emergency responders in finding useful information during crises, primarily focused on analysing the text of posted messages~\parencite{RN36, RN53, RN65}.}

\looseness -1 \ric{However, more recently there has been growing interest in the value-add of posted photos and other imagery during an emergency~\parencite{RN73, said2019natural, DBLP:conf/iscram/BuntainMS22}. Some papers aim to improve effectiveness of image classification on crisis imagery contents~\parencite{DBLP:conf/iscram/0002AQPO20, DBLP:conf/iscram/AsamiFHH22, DBLP:conf/iscram/LiCCIO19, DBLP:conf/iscram/LiC20,}. Indeed, some studies have shown that images posted on social media for events such as floods or wildfires can be useful when allocating resources or estimating damage severity~\parencite{nguyen2017damage, DMD, DBLP:conf/iscram/SoseaSCCR21, DBLP:conf/iscram/AkhtarO021}. As a result, the development of automated tooling to analyse crisis imagery and categorize it into useful types is of growing importance. To-date, deep convolutional neural networks, e.g., ResNet and VGG16~\parencite{crisisdbenchmark}, have been a popular solution for crisis image content categorization, that have reported high accuracy. These solutions rely on general pre-trained models produced from non-crisis image datasets as a starting point, e.g. ImageNet, and then fine-tune those models for a downstream task via transfer learning. It is not obvious why a model initially trained to identify mundane objects like cats or buildings would be effective for identifying images of people needing to be rescued. Moreover, recent advances in the field of computer vision have introduced alternative transformer-based neural architectures~\parencite{RN83}, which are suitable for large-scale multi-task pre-training.}

Hence, in this paper, we examine whether we can improve the performance of crisis image classification tasks via models pre-trained using in-domain crisis imagery, rather than relying on a general image classification model as a starting point. In particular, using the state-of-the-art ViT architecture~\parencite{RN83} as a base we pre-train new models using crisis imagery from the new incidents1M image dataset~\parencite{incident1m}, which we refer to as CrisisViT models. We have released these models for the community, and they can be downloaded via:
\begin{itemize}
    \item \url{https://github.com/longkukuhi/CrisisViT}
\end{itemize}
Through experimentation over the Crisis Image Benchmark dataset~\parencite{crisisdbenchmark}, we show that CrisisViT is more effective than previous state-of-the-art deep convolutional neural models, with an increase in accuracy of 3.90 absolute points (from 79.18\% to 83.07\%).
Moreover, the proposed best CrisisViT outperforms all baselines, as well as an existing ViT model, by up to 1.25\% absolute accuracy on average. This demonstrates that a dedicated large-scale crisis image dataset is key for the crisis image content categorization task.

\section{Related Work}

\ric{\noindent \textbf{Image content from Social Media platforms for Crisis Response}: Social media is increasingly seen as a critical information and communication platform during emergencies, as a channel to gather and analyze urgent information during a crisis~\parencite{RN84, RN85, RN86}. However, the majority of prior work in this space has focused on analysing textual content posted to these platforms rather than imagery~\parencite{RN73}.  On the other hand, recent works have begun to explore the value-add that crisis images posted to social media platforms can bring, as well as how to minimise the costs associated to image analysis through AI automation. For example, \cite{nguyen2017damage} demonstrated that crisis images on social media can be used for a variety of humanitarian aid activities (such as identifying areas in need of goods and services). Meanwhile, \cite{alam2017image4act} showed that social media images are helpful for damage assessment during flooding events, while \cite{daly2016mining} illustrated that geotagged images can be used to identify affected regions in need of aid. Functionally, crisis image analysis can be seen as a classification or tagging problem, where a human or machine needs to analyse the image and then assign a label or labels to that image. To-date the crisis image domain has largely focused on four image classification use-cases:}
\begin{itemize}
\item \ric{\emph{Disaster Type Detection}: The high-level classification of the type of disaster depicted within an image, such as an earthquake, fire or flood.} 
 \item \ric{\emph{Informativeness/Usefulness}: The classification of images to determine whether it contains some form of valuable information for an emergency responder. Typically represented as a binary informative/not informative classification.} 
\item \ric{\emph{Humanitarian Categories}: This form of classification is focused on what is happening within the image, where the goal is to identify images that are relevant to different types of humanitarian response activities. Common humanitarian categories include images of affected individuals, images of infrastructure or utility damage, or images of people needing rescue. }
\item \ric{\emph{Damage Severity}: Finally, one common use for crisis images is to judge the severity of damage in a particular area, which is useful for response prioritization or damage costing purposes. The damage severity task mainly targets three levels: severe damage, mild damage, and little or no damage.}
\end{itemize}
\ric{Notably, \cite{crisisdbenchmark} developed a standard dataset that combines training and test examples for all four tasks, which we use later in this paper to evaluate our models. }
%These works have shown that deep learning image models for identifying actionable information from imagery content within social media are feasible and to some degree, effective, although more work is still needed in this area~\parencite{RN65}.

%\looseness -1 Therefore, this paper focuses on improving the effectiveness of crisis imagery content categorization on social media through building a better image model, i.e. using a better base model and utilising the large-scale crisis image dataset. Therefore, it is first worth discussing the state-of-the-art models in the categorization of image data:

\ric{\noindent \textbf{Deep Neural Networks for Image Classification}:
In the wider field of image classification, the most dominant type of solution is the deep learned AI model. These approaches function by taking an embedding of the pixel data from the image as input, which is fed into a deep neural network that extracts some meaning from that image. A deep neural model is trained by example, where an image is provided to the model, the model then generates a predicted label, and then depending on whether it got the label correct feedback is transferred backward into the model, updating the network. Over the course of seeing thousands to millions of example images, the model learns what pixel embeddings correlate with the desired labels. However, deep neural networks are computationally expensive to train, and tend to exhibit higher accuracy if pre-trained on multiple related tasks~\parencite{rcnn,fasterrcnn,yolo}. Hence it is common for companies and researchers to release pre-trained neural models, which other developers can then adapt at a lower cost to their own use-case (known as transfer learning). With regard to the structure of the neural model itself, there are currently two competing architectures: convolutional neural networks (CNNs); and transformers. CNNs have traditionally been the dominant neural network type used for image classification. Transformer architectures have been shown to be highly effective for text classification~\parencite{RN53}, but under-perform when adapted for images due to the markedly higher dimensionality (there are more pixels in an image than words in a sentence). Architecturally CNNs are advantaged here, as their convolutional structure forces them to find the parts of the image that matter and discard the rest, enabling them to better generalize to unseen examples.  As a result, pre-trained CNNs are popular choices as baselines, such as ResNet152~\parencite{RN36} and VGG~\parencite{vgg16}.}
 
\ric{\noindent \textbf{Advances in Image Transformers}:
Over the past 5 years, significant research efforts have been made to improve the effectiveness of transformer architectures~\parencite{RN52} for image classification. The issue with applying transformers for image classification is two-fold: 1) training transformers on images is much more costly in comparison to a CNN, as transformer training time rapidly scales with input dimensionality due to its attention mechanism; and 2) transformers when applied to images have been shown to not generalize well to unseen images, as they lack some of the inductive biases that are learned naturally by CNNs. Early approaches tried to reduce training costs by applying self-attention to only the local neighbourhood for each query pixel~\parencite{RN93}, or by applying attention to only small parts of the image~\parencite{RN94}. However, an important breakthrough occurred with the development of the ViT model~\parencite{RN83}, which was the first vision transformer model to efficiently apply attention globally with minimal modifications to the transformer architecture. In 2021, Masked Autoencoders (MAE)~\parencite{RN82} were proposed, which then further addressed the cost of training via the use of a high image masking strategy with an encoder-decoder self-supervised pre-training schema, which enables MAE to learn how to reconstruct the original image based on only partial observations of that image. This approach reduces the number of pixels that need to be fed into the transformer and is the best current solution for reducing training time. Similarly, SimMIM~\parencite{xie2022simmim} proposed the use of masked image modelling to pre-trained vision transformers but without a decoder. Meanwhile, Data2vec~\parencite{data2vec} introduced a teach-student mode to pretrain vision transformers by representation learning. 
These models are normally pre-trained based on large-scale image datasets like ImageNet~\cite{RN91}, and they can then be `fine-tuned’ with new examples to transfer the pre-trained knowledge into the target downstream tasks, a process that is referred to as transfer learning~\cite{torrey2010transfer}. Although vision transformers dominate in the computer vision domain, the transferability of vision transformers remain unclear. As pointed out in \cite{RN83}, lacking discernible learned inductive biases limits the performance of vision transformers to handle downstream tasks. This appears to be a core limitation of transformers that cannot be easily overcome, leading to works such as \cite{zhou2021convnets} performing expensive whole network fine-tuning (that requires a large in-domain training dataset) to adapt the pre-trained model. In this work, we aim to push the boundaries of crisis image classification performance using transformers, and hence we need such a large in-domain crisis dataset. In our subsequent experiments, we evaluate whether the newly released incidents1M dataset~\parencite{incident1m} is sufficient to enable effective vision transformer models for crisis image classification.}

%\subsection{Deep Multi-modal learning}

% \noindent \textbf{Image classification in Crisis domain}: 

\section{Methodology}\label{sec:method}

\ric{In this work, we investigate whether pre-training on a large-scale crisis image dataset can improve the performance of crisis classification tasks. We choose a state-of-the-art transformer-based image classification model, ViT~\parencite{RN83}, as our base model and propose a new CrisisViT variant, which surpasses other deep learning image models in performance and robustness for a range of crisis image classification tasks. We use incidents1M~\parencite{incident1m} as the large-scale crisis image dataset for training. We try out various ways to pretrain CrisisViT, such as different pre-training strategies, examples used, and training labels, based on the dataset characteristics. We discuss the implementation of these models below.
When building the CrisisViT models, there are two main decisions that need to be made: 1) determine the dataset used to train; and 2) decide how to train with that dataset.}

\noindent \textbf{Pretrain datasets}:  \ric{For the pre-training dataset, we have the option of using either a known effective general image classification dataset, or attempt training with a more specialised in-domain dataset. To represent a general image classification dataset we use the popular ImageNet-1k image collection, while for an in-domain dataset we experiment with the new incidents1M crisis image collection:}
\begin{itemize}
\item \textbf{ImageNet-1k}: \ric{The ImageNet Large Scale Visual Recognition Challenge (ILSVRC)~\parencite{imagenet} project provides more than 14 million human-annotated images that could be used to train deep neural image models. These images are labelled based on the objects depicted, which might be mundane objects like cell phones, animals, or structures. Using such a dataset to pretrain a neural image classification model directs it towards the presence of known objects when later used for a downstream classification task. For example, it is intuitive that if trying to identify images about wildfires, being able to identify a firetruck in the image would be useful. In order to compare our result with other works, we follow the settings of \cite{crisisdbenchmark} and \cite{mae} that use the ImageNet-1k subset of ILSVRC, which is its most commonly used component. ImageNet-1k has 1,000 object classes (types of objects) and contains 1,281,167 training images, 50,000 validation images and 100,000 test images. In the rest of the paper, we refer to this dataset as ImageNet-1k.}

\item \textbf{Incidents1M}: \ric{Incidents1M is a large-scale crisis dataset of images taken during natural disasters. This dataset contains annotated labels for 1,787,154 images with two main types: 43 incident categories (e.g. airplane accident, bicycle accident, car accident.); and 49 place categories (e.g. building outdoor, highway, forest, ocean, sky.). Of the 1,787,154 images, just over half (977,088) contain one positive label, i.e. they belong to at least one of the (43+49) categories. It is possible for images to have multiple labels (belong to multiple categories). Notably, \cite{incident1m} does not release the image files, instead providing URLs pointing to those image files for other researchers to download. As such, over time as online content gets deleted or becomes unavailable this dataset will shrink. We downloaded this dataset during the final quarter of 2022, and at that time we retrieved 1,226,943 of the images (68.7\%). This crawled subset has 671,506 images labeled positively to one or more of the incident type categories and 522,782 images labeled positively to one or more of the place categories.}
\end{itemize}

Meanwhile, for the ways of utilising mentioned pre-train datasets, we can use these datasets either in isolation or together:

\begin{itemize}
\item \textbf{ImageNet-1k + Incidents1M}: \ric{Under this setting we load the pre-trained weights from MAE~\parencite{mae}---a model prebuilt via self-supervised learning on ImageNet-1k---to perform supervised pre-training on the ImageNet-1k object labels to encode information about the thousand object classes into the model. We further augment this model using training examples from the Incidents1M dataset to teach the model how to identify crisis-related information. This forms a new base model that we can later be fine-tuned for different (crisis image classification) downstream tasks.}

\item \textbf{Incidents1M only}: \ric{In this setting, instead of starting from an existing model, we take a blank model and conduct both self-supervised and supervised training using the images and labels in the Incidents1M dataset. This should act similarly to the above base model, but will lack some of the more general object recognition capabilities. In our later experiments, we compare these base models to determine whether starting from a more general image classification model or using only in-domain training is sufficient.}

\end{itemize}

\noindent \textbf{Pretrain tasks}: \ric{Importantly, the Incidents1M dataset~\parencite{incident1m} supports two main crisis categorization tasks: 1) incident type classification with 43 incident types; and 2) place type classification with 49 place types. In effect, this means that we can pre-train our base model using some or all of these 43+49 image types.  We experiment with four ways to utilise these training images and labels in our experiments:}

\begin{itemize}
\item \textbf{Binary training}: \ric{As we have 43+49 labeled image types, one methodology for pre-training a base model would be to consider each of these 92 image types as a different binary classification task. We can then incrementally train the base model to classify each of these 92 types in sequence, thereby incrementally building up the model’s ability to identify different types of crisis content. We refer to this as Binary classification pre-training and use it as a baseline. However, we remark that this might not be the best training strategy, as lessons learned by the model when training during types seen early may be overwritten by later types (a phenomenon known as catastrophic forgetting).}

\item \textbf{Incident OR Places training}: \ric{The second training methodology that we might employ would be to instead combine the images and labels for only one of the Incidents1M tasks, i.e. the 43 incident categories or the 49 place categories, into a unified set of training examples. In this way, we can see which of the two Incidents1M tasks provides more useful information for enhancing our downstream tasks. In contrast to binary training, in this setting, we do not divide our training by image type and instead train all image types for our selected task concurrently. Also, since an image can have multiple labels, we want to avoid sharing images across categories, in this scenario if an image has multiple labels, we consider it as belonging to only the category denoted by the first listed label.}

\item \textbf{Dual (Incident+Places) training}: \ric{The final pre-training methodology we use is duel training, which is identical to Incident or Places training, with the exception that we combine both tasks, rather than building separate models for each of the two Incidents1M tasks.}

\item \textbf{Self-supervised training}: Following the self-supervised training method from Masked Autoencoders~\parencite{RN82}, the CrisisViT model is trained using a self-supervised approach in which it learns to predict missing patches of an image by masking out random portions of the input image and then reconstructing the masked image. By employing this technique, the CrisisViT model can extract meaningful representations from large amounts of unlabeled data, leading to improved performance on image classification tasks.

\end{itemize}

All variants of CrisisViT use the same architecture as the ViT-base model~\parencite{RN83}, but with different hyperparameters. If the pertaining datasets are ImageNet-1k plus Incidents1M, it means we load the pretrain weights from~\parencite{RN83}.

\section{Experimental setup}\label{sec:setup}

\noindent \textbf{Downstream (Target) Dataset}:
\zijunnew{To evaluate how effective the CrisisViT model variants are, we require a downstream or target dataset, which represents one or more meaningful crisis image classification tasks. As discussed previously, the most common uses for crisis imagery are disaster type detection, informativeness/usefulness classification, grouping images into humanitarian categories, and damage severity estimation. Crisis Image Benchmark~\parencite{crisisdbenchmark} is a composite test collection that aggregates several datasets together, including CrisisMMD~\parencite{RN87}, data from AIDR~\parencite{imran2014aidr} and the Damage Multimodal Dataset (DMD)~\parencite{DMD}. The dataset consists of labels for four tasks:
\begin{itemize}
    \item Task 1: Disaster type classification
    \begin{itemize}
        \item Earthquake
        \item Fire
        \item Flood
        \item Hurricane
        \item Landslide
        \item Other disaster type
        \item Not disaster
    \end{itemize}
    \item Task 2: Informativeness
    \begin{itemize}
        \item Informative
        \item Not informative
    \end{itemize}
    \item Task 3: Classification into humanitarian categories
    \begin{itemize}
        \item Affected, injured, or dead people
        \item Infrastructure and utility damage
        \item Rescue volunteering or donation effort
        \item Not humanitarian
    \end{itemize}
    \item Task 4: Classification into damage severity categories
    \begin{itemize}
        \item Severe damage
        \item Mild damage
        \item Little or none
    \end{itemize}
\end{itemize}
} 

\zijunnew{This crisis image benchmark provides both training and testing for the four tasks. We follow the same experimental setup for the training, validation, and testing splits as in the original paper~\parencite{crisisdbenchmark}.} 

\noindent \textbf{Metrics}: We evaluate the performance of CrisisViT models in terms of their classification accuracy. Note that all metrics are reported on the same test set of the Crisis Image Benchmark dataset. Each experiment is run at least three times, and we report the average of the results.

\noindent \textbf{Baselines}:
\ric{Our overall goal in this paper is to determine to what extent a large-scale crisis image dataset (IncidentM1 in this case) improves the performance of transformer-based image classification models when performing crisis content categorization, as well as to determine best practices during training. Hence, in our later experiments we will compare our CrisisViT model to a number of either popular or state-of-the-art image classification models fine-tuned and evaluated on the crisis image benchmark, but that do not have knowledge on the Incidents1M dataset:}

\begin{itemize}
\item \textbf{ResNet101}: \parencite{RN36} proposes ResNet, a convolutional neural network with a deep residual connection, which achieves very high accuracy on the ImageNet dataset. ResNet101 is a deeper variant of ResNet with 101 layers.
\item \textbf{EffiNet (b1)}: \parencite{abs-1905-11946} studies model scaling and identifies that carefully balancing network depth, width, and resolution can lead to better performance. EffiNet (b1) is the second smallest version of EffiNet, which achieves similar performance to ResNet101 but with a smaller size.
\item \textbf{VGG16}: A convolutional neural network proposed by~\parencite{vgg16} that performs well on a wide range of tasks. It has 16 convolutional layers.
\item \textbf{ViT-Base}: ViT~\parencite{RN83} is the first vision transformer model to efficiently apply attention globally with minimal modifications to the transformer architecture, achieving remarkable performance on various datasets. ViT-Base is the base version of ViT with 12 layers and 768 hidden size dimension.
\end{itemize}

\noindent \textbf{CrisisViT Parameters}:
\ric{As with all machine learning models, there are a number of hyper-parameters that can affect the performance of the resultant model. We pre-trained CrisisViT with self-supervised learning and supervised learning on the Incidents1M dataset by using the Adam optimiser, a batch size of 1024 and 128, respectively, and the ReLU activation function.} We also experimented with other batch sizes [32,64,128,256,512], which led to lower performance of ViT. We fixed the training epoch for self-supervised training on the Incidents1M dataset at 400, and separately tested the models with supervised training pre-training steps of 10 and 20 epochs on the same dataset. The performances for each experiment are reported separately.

% \usepackage{multirow}
\begin{table}[]
\resizebox{\textwidth}{!}{\begin{tabular}{|l|l||l|l|l|c||ccccc|c|}
\hline
\multirow{3}{*}{Model} & \multirow{3}{*}{Type} & \multicolumn{4}{c|}{Pre-Training} & \multicolumn{5}{c|}{Task} & Training \\ 
\cline{3-11} 
&& Self-Supervised & \multicolumn{3}{c|}{Supervised}  & \multicolumn{1}{l|}{Disaster} & \multicolumn{1}{l|}{Info}  & \multicolumn{1}{l|}{Human} & \multicolumn{1}{l|}{Damage} & \multicolumn{1}{l|}{} & Time \\ 
\cline{3-11} 
&& Dataset & Dataset & Methodology & Epochs & \multicolumn{1}{c|}{Test}     & \multicolumn{1}{c|}{Test}  & \multicolumn{1}{c|}{Test}  & \multicolumn{1}{c|}{Test}   & AVG   & (hours)               \\ \hline
                       \hline
ResNet101 & CNN             & None                             & ImageNet-1k                          & Multi-Class (1k)                           & 10                     & \multicolumn{1}{c|}{81.3}     & \multicolumn{1}{c|}{85.2}  & \multicolumn{1}{c|}{76.5}  & \multicolumn{1}{c|}{73.7}   & 79.175     &  N/A            \\ \hline
EffiNet   (b1) & CNN         & None                             & ImageNet-1k                          & Multi-Class (1k)                           & 10                     & \multicolumn{1}{c|}{81.6}     & \multicolumn{1}{c|}{86.3}  & \multicolumn{1}{c|}{76.5}  & \multicolumn{1}{c|}{75.8}   & 80.05      &  N/A            \\ \hline
VGG16  & CNN                 & None                             & ImageNet-1k                          & Multi-Class (1k)                           & 10                     & \multicolumn{1}{c|}{79.8}     & \multicolumn{1}{c|}{85.8}  & \multicolumn{1}{c|}{77.3}  & \multicolumn{1}{c|}{75.3}   & 79.55       &  N/A           \\ \hline
ViT-Base  & TF              & ImageNet-1k                               & ImageNet-1k                          & Multi-Class (1k)                           & 20                     & \multicolumn{1}{c|}{84.10}    & \multicolumn{1}{c|}{86.59} & \multicolumn{1}{c|}{79.43} & \multicolumn{1}{c|}{77.18}  & 81.82   &  N/A              \\ \hline
\hline

CrisisViT   & TF             & Incidents1M                       & Incidents1M                  & Multi-Class (Incident)                      & 10                     & \multicolumn{1}{c|}{84.91}    & \multicolumn{1}{c|}{86.85} & \multicolumn{1}{c|}{79.43} & \multicolumn{1}{c|}{77.96}  & 82.29   & 36              \\ \hline
CrisisViT     & TF           & Incidents1M                       & Incidents1M                  & Multi-Class (Incident)                      & 20                     & \multicolumn{1}{c|}{84.73}    & \multicolumn{1}{c|}{86.61} & \multicolumn{1}{c|}{79.60} & \multicolumn{1}{c|}{77.31}  & 82.06       &  420          \\ \hline
CrisisViT    & TF            & Incidents1M                       & Incidents1M                  & Multi-Class (Places)                         & 10                     & \multicolumn{1}{c|}{84.95}    & \multicolumn{1}{c|}{87.85} & \multicolumn{1}{c|}{80.16} & \multicolumn{1}{c|}{\textbf{78.75}}  & 82.93 $\ast$   &  420              \\ \hline
CrisisViT   & TF             & Incidents1M                       & Incidents1M                  & Multi-Class (Places)                         & 20                     & \multicolumn{1}{c|}{\textbf{85.26}}    & \multicolumn{1}{c|}{\textbf{87.97}} & \multicolumn{1}{c|}{80.34} & \multicolumn{1}{c|}{78.72}  & \textbf{ 83.07 $\ast$}   &  420              \\ \hline

CrisisViT    & TF            & Incidents1M                       & Incidents1M                  & Multi-Class (Incident+Places)               & 10                     & \multicolumn{1}{c|}{85.01}    & \multicolumn{1}{c|}{86.85} & \multicolumn{1}{c|}{79.60} & \multicolumn{1}{c|}{77.31}  & 82.19     &  430          \\ \hline
CrisisViT     & TF           & Incidents1M                       & Incidents1M                  & Multi-Class (Incident+Places)               & 20                     & \multicolumn{1}{c|}{84.57}    & \multicolumn{1}{c|}{86.69} & \multicolumn{1}{c|}{79.23} & \multicolumn{1}{c|}{77.41}  & 81.98    & 430             \\ \hline

\hline
CrisisViT   & TF             & ImageNet-1k                               & ImageNet-1k+Incidents1M               & Multi-Class (1k) + Multi-Class 
(Incident)                      & 10                     & \multicolumn{1}{c|}{84.88}    & \multicolumn{1}{c|}{87.17} & \multicolumn{1}{c|}{79.64} & \multicolumn{1}{c|}{78.29}  & 82.49  & 34               \\ \hline
CrisisViT   & TF             & ImageNet-1k                               & ImageNet-1k+Incidents1M               & Multi-Class (1k) + Multi-Class (Incident)                      & 20                     & \multicolumn{1}{c|}{85.23}    & \multicolumn{1}{c|}{87.08} & \multicolumn{1}{c|}{79.71} & \multicolumn{1}{c|}{78.47}  & 82.62 $\ast$   & 48             \\ \hline
CrisisViT    & TF            & ImageNet-1k                               & ImageNet-1k+Incidents1M               & Multi-Class (1k) + Multi-Class (Places)                          & 10                     & \multicolumn{1}{c|}{85.04}    & \multicolumn{1}{c|}{87.04} & \multicolumn{1}{c|}{80.15} & \multicolumn{1}{c|}{78.16}  & 82.60 $\ast$  & 34              \\ \hline
CrisisViT    & TF            & ImageNet-1k                               & ImageNet-1k+Incidents1M               &Multi-Class (1k) +  Multi-Class (Places)                            & 20                     & \multicolumn{1}{c|}{85.04}    & \multicolumn{1}{c|}{87.51} & \multicolumn{1}{c|}{79.88} & \multicolumn{1}{c|}{78.01}  & 82.61 $\ast$ & 48               \\ \hline

CrisisViT    & TF            & ImageNet-1k                               & ImageNet-1k+Incidents1M               & Multi-Class (1k) + Binary (Incident+Places)                         & 20                     & \multicolumn{1}{c|}{81.13}    & \multicolumn{1}{c|}{84.15} & \multicolumn{1}{c|}{75.60} & \multicolumn{1}{c|}{75.71}  & 79.14  & 460               \\ \hline
CrisisViT    & TF            & ImageNet-1k                               & ImageNet-1k+Incidents1M               & Multi-Class (1k) + Multi-Class (Incident+Places)               & 10                     & \multicolumn{1}{c|}{85.01}    & \multicolumn{1}{c|}{87.13} & \multicolumn{1}{c|}{\textbf{80.42}} & \multicolumn{1}{c|}{78.19}  & 82.69 $\ast$ & 36
              \\ \hline
CrisisViT    & TF            & ImageNet-1k                               & ImageNet-1k+Incidents1M               & Multi-Class (1k) + Multi-Class (Incident+Places)                & 20                     & \multicolumn{1}{c|}{84.95}    & \multicolumn{1}{c|}{86.92} & \multicolumn{1}{c|}{79.40} & \multicolumn{1}{c|}{77.70}  & 82.24  & 52
               \\ \hline

\end{tabular}}
\caption{Experimental result overall baselines and variants of CrisisViT model. We use $\ast$  to denote a significant difference between the performances of the ViT baseline and the proposed model, according to the paired t-test
with the Holm-Bonferroni correction for p $<$ 0.01.}
\label{tb:result_compare_table.png}
\end{table}

\section{Experimental results}\label{sec:results}

\ric{To evaluate what extent the Incidents1M large-scale crisis image dataset can increase the performance of the image classification models for a range of tasks, we divide our analysis into three primary research questions, based on the different ways that Incidents1M can be utilised:}

\begin{itemize}
    \item RQ1: \zijunnew{To what extent can transformer-based architectures outperform traditional convolutional neural networks (CNNs) in image classification tasks?}
    \item RQ2: \zijunnew{What are the optimal pre-training strategies for the CrisisViT model when pre-training on the Incidents1M dataset?}
    \item RQ3: \zijunnew{Does starting from a more general image classification model, such as ImageNet-1k, provide significant advantages over using only in-domain training with the Incidents1M dataset in terms of the performance and robustness of the CrisisViT model for crisis image classification tasks?}
\end{itemize}

In this section, we report the results comparing the performances between state-of-the-art deep convolutional neural image models, the transformer-based image model ViT, and different variants of our proposed CrisisViT model, produced for crisis content categorization on four tasks of the Crisis image benchmark dataset.

\subsection{RQ1: ViT vs. Convolutional neural baselines}
\ric{We begin by determining how well a transformer image classification architecture like ViT performs for the domain of crisis image classification. Since most prior works (as discussed in the related work) employed the convolutional neural network (CNN) architecture, we intend to understand if transformer-based models make a difference. To this end, we compare three CNN baselines, namely ResNet101, EffiNet (b1) and VGG16, with the best transformer architecture ViT. What differentiates the three cases is their training starting point, i.e., the base model. We train the base model for four downstream (target) tasks to produce corresponding four models. The first four rows of Table~\ref{tb:result_compare_table.png} report the performance of these models under the test set for each task.}

\ric{As shown in Table~\ref{tb:result_compare_table.png}, Vit outperforms the other three CNN-based models. %, ViT is the best performing of these models, which confirms observations made when comparing CNN and transformer architectures for other image classification domains. 
Specifically, the ViT transformer model appears to be primarily advantaged when used for Disaster Type classification, Humanitarian category classification, and Damage Severity estimation, with reported gains over the next best CNN-based model by 3.5\%, 1.4\% and 1.4\% absolute classification accuracy, respectively. Meanwhile, a much smaller but notable gain of 0.3\% is observed for the Informativeness classification. Overall, this confirms our expectation that transformer-based models are the current state-of-the-art in this domain, and as such, we use ViT as our primary comparison point in the remainder of this paper.}

\subsection{RQ2: Pre-training using Incident Types and Place Categories}
\ric{Having shown that the Transformer architecture ViT is superior to prior CNN-based architectures and quantified our baseline performance, we now examine the core question of this work: can we further boost the performance using a large-scale crisis image dataset? The underlying rationale is that by teaching a deep neural model how to tackle a different but related task helps the model learn the downstream task better. In this scenario, we used the Incidents1M dataset with labels for 43 incident-type and 49 place categories. These tasks are different but conceptually related to our four downstream tasks (i.e., disaster type classification, informativeness/Usefulness, humanitarian category classification, and damage severity estimation). %We want to know if teaching the model about incident types and place identification makes training on the downstream tasks easier. 
Rows 5-10 of Table~\ref{tb:result_compare_table.png} report the accuracy of the ViT architecture when pre-trained using Incidents1M instead of ImageNet-1k (as was used in the ViT-Base baseline). We report performance when pre-training with Incident type labels (Multi-Class (Incident)), place categorization labels (Multi-Class (Place)), and both (Multi-Class (Incident+Places)). We also report performances for 10 and 20 epochs to illustrate how performance improves with more training time.}

\ric{From Table~\ref{tb:result_compare_table.png}, we observe that in nearly all cases pre-training on Incidents1M leads to superior performances for crisis image classification than using only ImageNet-1k. For instance, when pre-training with the Incidents1M place category labels for 20 epochs, we observe a statistically significant (p $\le$ 0.05) accuracy gain over ViT-Base of 1.16\%. Second, comparing the Incident and Place labels provided by Incidents1M, the Place labels result in the best-performing downstream models in all cases, while the Incident labels provide a comparably smaller benefit (and in one case it harms the performance). Furthermore, we notice that when combining the Incident and Place labels together, performance does not improve over using the Place labels alone, indicating that the Incident labels are redundant when the Place labels are available. Third, comparing the effect of providing more training, when pre-training using the place labels, more training time (20 epochs) does lead to small performance gains (around 0.1-0.3\%). However, when providing additional training time to the Incident labels, downstream accuracy tends to degrade, confirming that teaching the model about the Incidents1M incident types leads to questionable benefits. Overall, we can conclude that pre-training with an in-domain dataset can lead to performance gains. Indeed, we observed between 0.9\% and 1.5\% accuracy across the downstream tasks tested. On the other hand, the inconsistent performance of the models pre-trained with the Incident labels indicates that not all conceptually related tasks are useful evidence for pre-training, and so researchers and developers should be selective regarding what datasets to use for pre-training.}

\subsection{RQ3: ImageNet-1k + Incidents1M?}
\ric{In the previous set of experiments we replaced ImageNet-1k-based pre-training with pre-training using the in-domain dataset i.e., Incidents1M. However, given that ImageNet-1k forms the basis for many effective image classification models in the literature, as well as providing a strong baseline (via ViT-Base), it is worth investigating whether we can instead augment ImageNet-1k training instead of replacing it. Hence, we pre-train a second set of base models that take the same base model as ViT-Base subject to ImageNet-1k self-supervised and supervised training and then add further pre-training using the Incidents1M data. If ImageNet-1k provides value on top of Incidents1M, combining both should result in a small improvement in accuracy on the downstream tasks. The final seven rows of Table~\ref{tb:result_compare_table.png} report classification accuracy on the four downstream tasks when we combine ImageNet-1k and Incidents1M pre-training.} 

\ric{In Table~\ref{tb:result_compare_table.png} we can observe that the best performing Incidents1M models pre-trained on the Places labels does not show performance improvements despite %, we observe that the majority of the time, 
providing additional ImageNet-1k training. %does not result in a performance uplift. 
This might lead us to conclude that ImageNet-1k is not adding value. However, if we investigate both the Incidents1M models that are pre-trained on the Incident or Incident+Places labels, we do observe a small performance uplift for the majority of tasks. Indeed, one of these models achieved the best overall performance for the Humanitarian categorisation task. On the other hand, given the small degree of the performance difference, it is not apparent that starting from the pre-trained ViT-Base model is better than training a new model. %from the first principles.
}

\section{Conclusions}\label{sec:conclusion}

\zijunnew{Social media has become an increasingly important platform for emergency response agencies to obtain valuable information, particularly images, for various crisis response tasks. However, due to the sheer volume of content on social media, automated techniques for filtering and classifying images are necessary. Existing methods rely on convolutional neural networks pre-trained on the general ImageNet-1k dataset, but recent developments in transformer-based image classifiers in combination with increased availability of tagged crisis imagery (via the Incidents1M dataset) have opened up new possibilities. In this paper, we introduced CrisisViT, a transformer-based architecture pre-trained on the Incidents1M dataset, which can be adapted for multiple downstream crisis image classification tasks. Through experimentation on the four tasks (Disaster Type classification, Informativeness classification, Humanitarian Category classification, and Damage Severity estimation) supported by the Crisis Image Benchmark dataset, we show that pre-training on the Incidents1M dataset can lead to significant improvements in accuracy, with an average absolute gain of 1.25\% over the four crisis image classification tasks tested. This work represents an important step towards building more effective crisis response tools that can utilize social media image data to support emergency response efforts.}

\printbibliography[heading=bibliography]

\end{document}